%% file: main_arxiv.tex
\documentclass[sigconf,nonacm,screen]{acmart}

\usepackage{soul} %

\usepackage{subcaption}
\usepackage{multirow}
\usepackage{bm}
\usepackage{ifthen}

\usepackage{pifont}

\usepackage{algorithm}
\usepackage{listings}
\usepackage[british,american]{babel}

\usepackage{xcolor}
\definecolor{tablegray}{gray}{0.45}

\usepackage{tabulary}
\newcolumntype{x}[1]{>{\centering\arraybackslash}p{#1pt}}
\newcommand{\tablestyle}[2]{\setlength{\tabcolsep}{#1}\renewcommand{\arraystretch}{#2}\centering\footnotesize}
\newlength\savewidth\newcommand\shline{\noalign{\global\savewidth\arrayrulewidth
		\global\arrayrulewidth 1pt}\hline\noalign{\global\arrayrulewidth\savewidth}}

\usepackage{colortbl}
\definecolor{Gray}{gray}{.9}

\usepackage{etoolbox}
\makeatletter
\AfterEndEnvironment{algorithm}{\let\@algcomment\relax}
\AtEndEnvironment{algorithm}{\kern2pt\hrule\relax\vskip3pt\@algcomment}
\let\@algcomment\relax
\newcommand\algcomment[1]{\def\@algcomment{\footnotesize#1}}
\renewcommand\fs@ruled{\def\@fs@cfont{\bfseries}\let\@fs@capt\floatc@ruled
	\def\@fs@pre{\hrule height.8pt depth0pt \kern2pt}%
	\def\@fs@post{}%
	\def\@fs@mid{\kern2pt\hrule\kern2pt}%
	\let\@fs@iftopcapt\iftrue}
\makeatother

\def\mathbi#1{\textbf{\em #1}}

\definecolor{Highlight}{HTML}{39b54a}  %
\definecolor{red}{HTML}{ff3509}  %
\renewcommand{\hl}[1]{\textcolor{Highlight}{#1}}

\usepackage{tcolorbox}

\DeclareRobustCommand\onedot{\futurelet\@let@token\@onedot}
\def\@onedot{\ifx\@let@token.\else.\null\fi\xspace}
\def\eg{\emph{e.g}\onedot} 
\def\ie{\emph{i.e}\onedot}

\def\etal{\emph{et al}\onedot}

\def\eg{\emph{e.g.}}
\def\ie{\emph{i.e.}}

\def\etal{\emph{et al.}}

\AtBeginDocument{%
  \providecommand\BibTeX{{%
    \normalfont B\kern-0.5em{\scshape i\kern-0.25em b}\kern-0.8em\TeX}}}

\setcopyright{acmcopyright}
\copyrightyear{2022}
\acmYear{2022}
\acmDOI{XXXXXXX.XXXXXXX}

\acmSubmissionID{701}

\begin{document}

\title{RepSR: Training Efficient VGG-style Super-Resolution Networks with Structural Re-Parameterization and Batch Normalization}

\vspace{-0.4cm}

\author{Xintao Wang}
\affiliation{%
  \country{ARC Lab, Tencent PCG}}
\email{xintao.alpha@gmail.com}

\author{Chao Dong}
\affiliation{%
 \country{Shenzhen Institute of Advanced Technology, Chinese Academy of Sciences; Shanghai AI Laboratory}}
\email{chao.dong@siat.ac.cn}

\author{Ying Shan}
\affiliation{%
  \country{ARC Lab, Tencent PCG}}
\email{yingsshan@tencent.com}

\begin{abstract}
This paper explores training efficient VGG-style super-resolution (SR) networks with the structural re-parameterization technique. The general pipeline of re-parameterization is to train networks with multi-branch topology first, and then merge them into standard $3{\times}3$ convolutions for efficient inference. In this work, we revisit those primary designs and investigate essential components for re-parameterizing SR networks. First of all, we find that batch normalization (BN) is important to bring training non-linearity and improve the final performance. However, BN is typically ignored in SR, as it usually degrades the performance and introduces unpleasant artifacts. 
We carefully analyze the cause of BN issue and then propose a straightforward yet effective solution. In particular, we first train SR networks with mini-batch statistics as usual, and then switch to using population statistics at the later training period. While we have successfully re-introduced BN into SR, we further design a new re-parameterizable block tailored for SR, namely RepSR. It consists of a clean residual path and two expand-and-squeeze convolution paths with the modified BN. Extensive experiments demonstrate that our simple RepSR is capable of achieving superior performance to previous SR re-parameterization methods among different model sizes. 
In addition, our RepSR can achieve a better trade-off between performance and actual running time (throughput) than previous SR methods.
Codes will be available at \url{https://github.com/TencentARC/RepSR}.
\end{abstract}



\maketitle

\input{sections/1_introduction}

\input{sections/2_related_work}
\input{sections/3_method}

\input{sections/4_experiments}

\section{Conclusion}
We have proposed RepSR, a simple yet effective re-parameterizable block tailored for SR.
In order to introduce the BN training non-linearity, we carefully analyze the cause of BN artifacts and propose an effective strategy to address this BN issue. 
Delicate designs such as clean residual path and expand-and-squeeze convolutions, are employed to further improve the re-parameterization performance. 
The proposed RepSR allows us to train efficient VGG-style SR networks, achieving better performance with faster inference speed.

\bibliographystyle{ACM-Reference-Format}
\bibliography{bib}

\end{document}

%% file: sections/1_introduction.tex
\section{Introduction}

Image super-resolution (SR) aims at recovering high-resolution (HR) images from their low-resolution (LR) counterparts.
While recent deep-learning-based methods~\cite{dong2016image,ledig2017srgan,lim2017edsr} have achieved excellent restoration performance, the inference speed is still a key challenge for real-world SR applications~\cite{zhang2020aimefficientsr,zhang2019aimconstrainedsr}. 
Though many compact network designs with fewer FLOPs (floating-point operations) have been proposed to meet this challenge, they may not run faster in real applications, as FLOPs cannot precisely reflect the actual speed. 
The widely-used skip and dense connections will also introduce extra costs on memory consumption. 
Instead, plain networks with a stack of  $3{\times}3$ convolutions and activation functions (\ie, VGG-style networks) can better take advantage of the GPU parallelization and hardware acceleration. Therefore, such plain structures (\eg, FSRCNN~\cite{dong2016accelerating}, ESPCN~\cite{caballero2017real}) are still popular in practical applications.

\begin{figure}[t]
	\begin{center}
		\includegraphics[width=\linewidth]{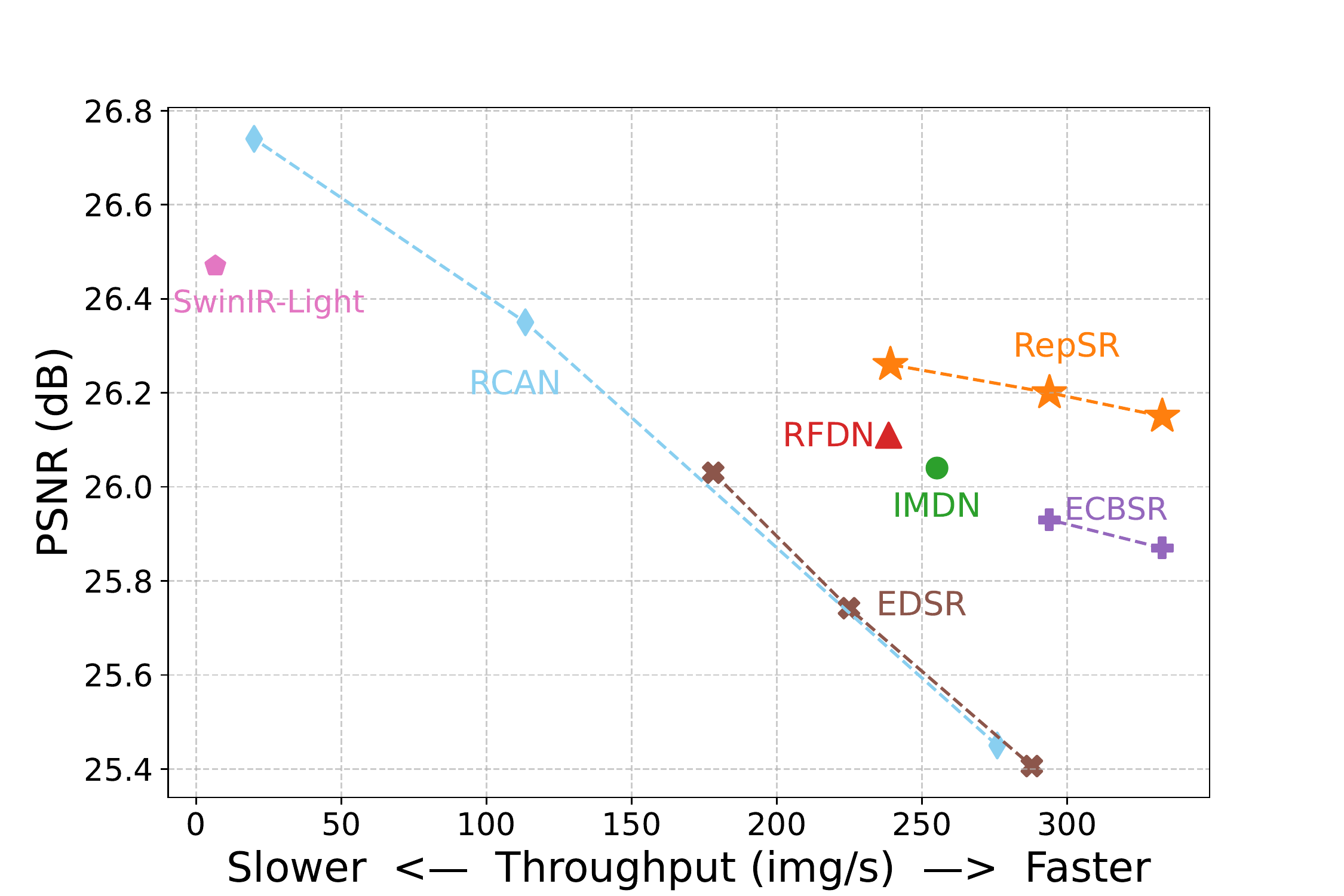}
		\caption{Comparisons of the trade-off between PSNR performance and the throughput. The throughput is tested on A100 with a batch is 4, fp16 precision and upsampling from $320\times180$ to $1280\times720$. Evaluated on the Urban100 test dataset~\cite{huang2015single} with the Y channel. We compare our RepSR with SwinIR-Light~\cite{liang2021swinir}, RCAN~\cite{zhang2018rcan}, EDSR~\cite{lim2017edsr}, ECBSR~\cite{zhang2021ecbsr}, IMDN~\cite{hui2019imdn} and RFDN~\cite{liu2020residual}.}
		\label{fig:teaser}
	\end{center}
\end{figure}

Directly training plain networks usually leads to inferior performance to multi-branch architectures (\eg, ResNet~\cite{he2016resnet}, EDSR~\cite{lim2017edsr}), mainly due to the inefficient optimization issues.
Re-parameterization \cite{zagoruyko2017diracnets,ding2021repvgg} is thus proposed to use a multi-branch architecture during training and switch to a plain network for testing. 
Specifically, it trains models with a linear combination of multiple branches and merges them into a standard $3{\times}3$ convolution in the inference stage.
Then the training effectiveness improves but with no sacrifice of inference time. 
To fully make use of the multi-branch topology,  it is essential to add \textit{batch normalization (BN)}~\cite{ioffe2015batch} in each branch. BN can introduce non-linearity during training and improve the final performance. Other activation functions can hardly take the place of BN, as they cannot be merged into a single convolution. The necessity of BN is also highlighted in RepVGG ~\cite{ding2021repvgg}.

However, BN seems to be unfriendly to the SR task. Previous literature usually suggests removing BN in the network structures~\cite{lim2017edsr,wang2018esrgan}, as BN not only decreases the performance~\cite{lim2017edsr}, but also brings annoying artifacts~\cite{wang2018esrgan}.
Zhang \etal~\cite{zhang2021ecbsr} have tried to directly employ existing re-parameterization blocks~\cite{ding2021repvgg} without BN in SR, but find no apparent improvement.
In order to make up for the loss caused by the absence of BN, they resort to designing more complicated blocks that can be re-parameterized. They propose an edge-oriented convolution block (ECB), which consists of classical Sobel and Laplacian filters~\cite{sun2008image}. 
Such an artificial design can only work well on tiny networks, but cannot generalize to larger ones in our experiments. 
Moreover, training ECB is slow, as it requires many additional depth-wise convolutions to incorporate those pre-defined filters. 

In this work, we aim to directly address the BN issue in SR, and propose a simple yet effective re-parameterizable block with the BN training non-linearity.
First, we carefully analyze the performance drop and artifacts caused by BN, and attribute this issue to the train-test inconsistency. 
Such inconsistency happens at the ``\textit{patch level}'' instead of ``dataset level'', which is contrary to our common sense.
Interestingly, our analysis also shows that this BN issue only appears during inference, while the training process is stable and healthy.
Intuitively, this inconsistency could be eliminated by training with population statistics.
Therefore, we propose to first train SR networks with mini-batch statistics, as normal BN does, and then switch to using population statistics at the later training period. 
In such a way, we can enjoy the benefits from faster convergence with BN training, while not being affected by BN artifacts.
\footnote{This strategy is named FrozenBN in high-level tasks, which is usually for transferring a pretrained model to downstream tasks or for small-batch-size training.
More discussions are in Sec.~\ref{subsec:repsr} and Sec.~\ref{sec:bn_strategy}.}

After re-introducing BN into SR, we then design a re-parameterizable block tailored for SR, namely, RepSR.
To be specific, 
1) a clean residual path is employed in RepSR.
The clean shortcut connection could better propagate detail information among blocks~\cite{he2016identity}.
2) We adopt channel expand-and-squeeze convolutions for other branches, \ie, $3{\times}3$Conv--BN--$1{\times}1$Conv. The over-parameterization~\cite{guo2018expandnets,arora2018optimization} typically improves the optimization efficiency, while wider features boost the network expressiveness~\cite{yu2020wide}. %
Our experiments show that the proposed RepSR with the improved BN is capable of achieving comparable or even superior performance than previous re-parameterization methods~\cite{zhang2021ecbsr,bhardwaj2021collapsible,ding2021repvgg} and is effective for both small and large models.

The contributions are summarized as follows.
\textbf{(1)} We provide a detailed analysis about the cause of BN artifacts, and propose a straightforward strategy to effectively address this issue. 
\textbf{(2)} A simple yet effective re-parameterizable block for SR is proposed. It achieves superior performance to previous re-parameterization methods among different model sizes.
\textbf{(3)} Equipped with RepSR, we are able to train efficient VGG-style SR networks. It can achieve a better trade-off between performance and throughput than previous SR methods.

%% file: sections/2_related_work.tex
\section{Related Work}
\noindent\textbf{The super-resolution} field~\cite{dong2014learning,ledig2017srgan,lim2017edsr,zhang2018learning,zhang2020deep,gu2019blind,luo2020unfolding,wang2021unsupervised,zhang2021designing,wang2021realesrgan} has witnessed a variety of developments since the pioneer work of SRCNN~\cite{dong2014learning,dong2016image}.
Different architectures are proposed, such as deeper networks~\cite{kim2016accurate,lai2017deep,haris2018deep}, residual blocks~\cite{ledig2017srgan,lim2017edsr,zhang2018residual,wang2018esrgan,wang2018sftgan}, recurrent architectures~\cite{kim2016deeply,tai2017image}, and attention mechanism~\cite{zhang2018rcan,dai2019second,liu2018non,mei2020image}.
Due to the limitation of computation resources and the challenge of heavy dense predictions in SR, efficient SR is urgent for real-world applications. 
CARN (cascading residual network)~\cite{ahn2018carn} with group convolutions is proposed to reduce the number of FLOPs. Hui \etal~\cite{hui2018idn,hui2019imdn} proposed an information multi-distillation network (IMDN) to compress the number of filters per layer. More designs can be found in the AIM challenge reports~\cite{zhang2020aimefficientsr,zhang2019aimconstrainedsr}.
However, fewer FLOPs and parameters do not necessarily mean higher efficiency~\cite{ding2021repvgg, zhang2021ecbsr}. 
The widely-used skip and dense connections also introduce extra costs on memory consumption. 
Instead, plain networks can better take advantage of the GPU parallelization and hardware acceleration. Therefore, such plain structures, \eg, FSRCNN~\cite{dong2016accelerating}, are still popular in practical applications. 
In this work, we aim to employ plain networks for efficient SR and propose a simple yet effective method to train such plain networks. 

\noindent\textbf{Batch normalization}~\cite{ioffe2015batch} is rarely used in SR, as researchers empirically found that 
BN degrades performance~\cite{lim2017edsr} and brings annoying artifacts~\cite{wang2018esrgan}. 
The strengths and weaknesses of BN have been deeply discussed in high-level tasks~\cite{wu2021rethinking}, while it still lacks the analysis in SR. We thus provide an in-depth investigation and introduce a straightforward way to address the BN issue in SR.
The benefit of switching to FrozenBN in the middle of training has also been observed in~\cite{wu2021rethinking,johnson2018image,xie2019intriguing}.

\noindent\textbf{Re-parameterization} is a recent technique to train plain networks~\cite{zagoruyko2017diracnets,ding2021repvgg}, boost the representative capacity of normal convolutions~\cite{ding2021ddb,ding2019acnet} or accelerate the network training~\cite{arora2018optimization}.
For SR re-parameterization, ECBSR~\cite{zhang2021ecbsr} introduces an edge-oriented convolution block, while SESR~\cite{bhardwaj2021collapsible} proposes a collapsible linear block.
But they simply discard BN layers together with training non-linearity, leading to a performance drop or more complicated designs. 
Instead, we re-introduce BN into SR and design a simple yet effective re-parameterizable block tailored for SR.

%% file: sections/3_method.tex
\section{Methodology}
\label{sec:methodology}

In this section, we describe our proposed RepSR in details, including the neat base network, the RepSR block and the re-parameterization for efficient inference.
The analysis of BN artifacts  in super-resolution and the proposed strategy for removing such artifacts are in Sec.~\ref{sec:revisit_bn}.

\subsection{Base Network}
\label{sec:base_network}
\begin{figure*}[t]
	\begin{center}
		\includegraphics[width=0.9\linewidth]{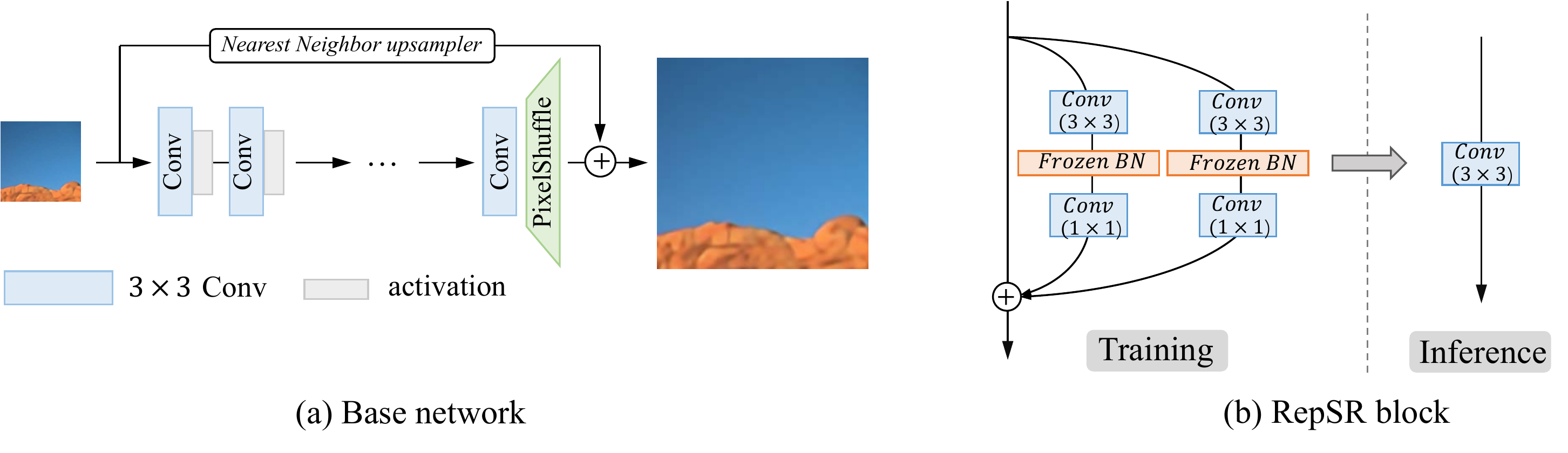}
		\caption{(a) The neat base network. It has a plain VGG-style network structure with a stack of $3{\times}3$ convolutions and activation layers. (b) RepSR block. It consists of a clean residual path, two expand-and-squeeze paths with Frozen BN layers. During inference, the RepSR block can be merged into one standard $3{\times}3$ convolution.}
		\label{fig:blocks}
	\end{center}
\end{figure*}

We first present the neat base network for efficient SR, as shown in Fig.~\ref{fig:blocks} (a).
In order to keep low computation cost and memory consumption, the base network has a plain VGG-style network structure, \ie, using a stack of $3{\times}3$ convolutions and activation layers.
We perform the upsampling operation at the end of network with a pixel-shuffle~\cite{shi2016real} layer.
A cheap nearest neighbor upsampler is adopted as the global skip connection.
We employ PReLU~\cite{he2015delving} as the activation function in our experiments.

Such a neat base network is also used in~\cite{zhang2021ecbsr} and has proven to be more deployment-friendly than other complicate network structures.
For example, as measured in~\cite{zhang2021ecbsr}, FSRCNN~\cite{dong2016accelerating} with a plain network structure has higher FLOPs than IMDN-RTC~\cite{hui2019lightweight} with a complicated topology, but runs about two-order faster on SnapDragon 865 for 540p to 1080p upscaling.

\subsection{Re-parameterization Block for SR}
\label{subsec:repsr}
Directly training plain VGG-style networks with $3{\times}3$ convolutions usually leads to inferior performance to multi-branch architectures (\eg, EDSR with residual paths~\cite{lim2017edsr}). 
Re-parameterization~\cite{zagoruyko2017diracnets,ding2021repvgg} is thus proposed to use a multi-branch architecture during training and then merge them into standard $3{\times}3$ convolutions for testing.
In their architectures, batch normalization (BN) is essential to introduce non-linearity during training and improve the final performance~\cite{ding2021repvgg}. 

In super-resolution, BN is usually removed from the network structures~\cite{lim2017edsr,wang2018esrgan}, as BN not only decreases the performance~\cite{lim2017edsr}, but also brings annoying artifacts.
In order to  make up for the loss caused by the absence of BN,
ECBSR~\cite{zhang2021ecbsr} designs complicated blocks that consists of classical Sobel and Laplacian filters~\cite{sun2008image}.

Instead, we propose the RepSR block, which re-introduces BN into re-parameterization in SR. We further improve the detailed structures of RepSR, so that it is more suitable for the SR task.
Structure simplicity is also considered in the design of RepSR.

\noindent\textbf{RepSR block.}
The overview of our proposed RepSR block is depicted in Fig.~\ref{fig:blocks} (b).
It consists of a clean residual path, two expand-and-squeeze paths with BN layers.

\textbf{1. Clean residual path.}
We employ a clean residual path without BN in RepSR, which is different from the design in RepVGG~\cite{ding2021repvgg}.
In RepVGG~\cite{ding2021repvgg} for high-level tasks, a residual path with BN is important. 
Experiments show that our clean residual path achieves better performance than a residual path with BN. 
It is mainly due to the different information processing mechanisms between SR and high-level tasks. 
In SR, the latter block \textit{enhances} the features from the former block, rather than converting them into more semantic spaces as those high-level tasks do.
Thus, a clean shortcut connection could better propagate detail information among blocks~\cite{he2016identity}.

\textbf{2. Expand-and-squeeze convolution.}
We adopt the channel expand-and-squeeze convolution for the other two branches~\cite{zhang2021ecbsr,ding2021ddb}.
Specifically, we first use a $3{\times}3$ convolution to expand channel dimension and then employ a $1{\times}1$ convolution to squeeze back the channel number, \ie, $3{\times}3$Conv--BN--$1{\times}1$Conv.

The output features $\mathbi{O}$ can be obtained from the input features $\mathbi{I}$ by:
\begin{align}
	\mathbi{O} &= \mathtt{BN}[(\mathbi{I} \circledast \mathbi{W}_3 + \mathbi{b}_3)] \circledast \mathbi{W}_1 + \mathbi{b}_1 
\end{align}
where $\mathbi{W}_3$, $\mathbi{b}_3$ are the weight and bias of $3{\times}3$ conv;  $\mathbi{W}_1$, $\mathbi{b}_1$ are the weight and bias of $1{\times}1$ conv.
This design is based on the two observations from previous works:
1) the over-parameterization~\cite{guo2018expandnets,arora2018optimization} typically improves the optimization efficiency;
2) wider features boost the network expressiveness~\cite{yu2020wide}. %
Between the two convolutions, BN is used for training non-linearity, and also for faster training convergence.

\textbf{3. Frozen BN layer.}
In this work, we carefully analyze the BN in SR and re-introduce BN into SR under the context of re-paramerization. 
We propose to first train SR networks with mini-batch statistics,
as normal BN does, and then switch to using population
statistics at the later training period. 
This strategy is named \textit{FrozenBN} in high-level tasks.
Although we use the same terminology, they are different in three aspects. 
\textbf{1)} Frozen BN is typically used in high level tasks, while we provide the first analysis and application in the SR task.
\textbf{2)} Frozen BN is usually used to transfer a pretrained model to downstream tasks or for small-batchsize
training. Instead, we adopt it for the training non-linearity in SR re-paramerization.
\textbf{3)} Different from previous frozen BN that is frozen for most of the training period, our proposed BN is trained for most of the training period, and is only frozen for a very short period (\eg, <$10\%$) at the end of training.
The analysis of BN artifacts in super-resolution and the proposed strategy for removing such artifacts are in Sec.~\ref{sec:revisit_bn}.

\textbf{Differences to ECBSR~\cite{zhang2021ecbsr}.}
ECBSR is the first re-parameterization work in SR for efficient inference.
\textbf{1)} Our RepSR has a simple yet effective structure for re-parameterization, indicating that the complicate Sobel and Laplacian filters in ECBSR are not necessary. This is mainly because we re-introduce the BN into SR, so that RepSR enjoys the non-linearity benefits from BN during training, while not being affected by BN artifacts in SR.
\textbf{2)} The performance of ECBSR degrades drastically for deeper networks (\eg, > 16 convolution layers) due to the complicated structure and optimization issues. While our RepSR can achieve good performance for both tiny and large networks. 
\textbf{3)} Due to the simplicity design, our RepSR has a faster training speed than ECBSR. For example, it can reduce the training time by $50\%$.

\textbf{Differences to RepVGG~\cite{ding2021repvgg}.}
RepVGG achieves excellent re-parameterization perforamnce in high level tasks.
\textbf{1)} Directly applying RepVGG block in SR cannot achieve satisfactory performace, as the BN decreases the performance and introduces unpleasant artifacts. To tackle the BN problem, we provide a detailed analysis of BN artifacts and propose the Frozen BN strategy for SR re-parameterization. 
\textbf{2)} We further improve the block designs that are tailed for the SR tasks, \eg, clean residual path, expand-and-squeeze convolutions.
 With such delicate designs, RepSR can achieve better performance than RepVGG.

\begin{figure}[t]
\begin{algorithm}[H]
\caption{Pseudocode for Re-parameterization}
\label{alg:code}
\algcomment{
	\textbf{Notes}:  \texttt{mm} (\texttt{bmm}) is (batch) matrix multiplication. \texttt{Reshape} here also includes \texttt{permute} operation if necessary.
}

\definecolor{codeblue}{rgb}{0.25,0.5,0.5}
\definecolor{codekw}{rgb}{0.1, 0.1, 0.1}
\footnotesize
\hspace{-4.06cm}\textbf{Step 1: Merge BN to its preceding conv} \\
\hspace{-4cm}\textbf{Step 2: Merge K${\times}$K Conv and $1{\times}1$ Conv}
\begin{lstlisting}[language=python]
  # K: kernel size, here equals to 3
  # c3w, c3b: weight and bias for the 3x3 convolution
  # c1w, c1b: weight and bias for the 1x1 convolution
  # Cout, Cmid, Cin: output, middle, input channel number

  # merge weight
  Reshape c3w from (Cmid, Cin, K, K) to (K^2, Cmid, Cin)
  Reshape c1w from (Cout, Cmid, 1, 1) to (1, Cout, Cmid)
  merged_weight = bmm(c1w.expand(K^2, Cout, Cmid), c3w)
  Reshape merged_weight from (K^2, Cout, Cin) to (Cout, Cin, K, K)

  # merge bias
  Reshape c3b from (Cmid, ) to (Cmid, 1)
  Reshape c1w from (Cout, Cmid, 1, 1) to (Cout, Cmid)
  merged_bias = mm(c1w, c3b).view(-1,) + c1b
\end{lstlisting}
\hspace{-5.2cm}\textbf{Step 3: Merge residual path}
\definecolor{codekw}{rgb}{0.85, 0.18, 0.50}
\begin{lstlisting}[language=python]
  for i in range(Cout):
      merged_weight[i, i, 1, 1] += 1
  # There are no changes to bias
\end{lstlisting}
\end{algorithm}
\end{figure}

\subsection{Re-parameterization for efficient inference}

We now describe how to merge the RepSR block into one standard $3{\times}3$ convolution.
The overall procedure is listed in Algorithm~\ref{alg:code}.

We first merge BN into its preceding convolution. This procedure follows previous practice and is easy to implement.
Then, we merge the $3{\times}3$ conv and $1{\times}1$ conv. Formally, the output features $\mathbi{O}$ can be obtained from the input features $\mathbi{I}$ through the following formula:
\begin{align}%
	\label{eq:merge:2} 
	\mathbi{O} &= (\mathbi{I} \circledast \mathbi{W}_3 + \mathbi{b}_3) \circledast \mathbi{W}_1 + \mathbi{b}_1 \\
	\label{eq:merge:3}
	  &=\mathbi{I} \circledast \underbrace{(\mathbi{W}_3 \circledast \mathbi{W}_1)}_{\text{merged weight}} + \underbrace{(\mathbi{b}_3 \circledast \mathbi{W}_1 + \mathbi{b}_1)}_{\text{merged bias}} 
\end{align}
where $\mathbi{W}_3$, $\mathbi{b}_3$ are the weight and bias of $3{\times}3$ conv;  $\mathbi{W}_1$, $\mathbi{b}_1$ are the weight and bias of $1{\times}1$ conv.
According to the associativity of convolution, we can obtain the merged weight and bias from Eq.~\ref{eq:merge:2}.
As the second convolution is a $1{\times}1$ convolution, which performs only channel-wise
linear combination without spatial aggregation, the merge can be easily conducted by (batch) matrix multiplication, as show in Algorithm~\ref{alg:code}.
As for the residual path, it adds one to the center of $3{\times}3$ kernel on the diagonal of the first two dimensions. There are no changes to bias in this case.

\noindent\textbf{The order of $3{\times}3$ Conv and $1{\times}1$ Conv}.
For the expand-and-squeeze convolution branch, previous works usually first adopt a $1{\times}1$Conv followed by a $3{\times}3$Conv~\cite{ding2021ddb,zhang2021ecbsr}.
However, such a sequential of convolutions cannot be directly re-parameterized due to the padding issue. 
Specifically,
Eq.~\ref{eq:merge:3} does not hold when $\mathbi{I}$ first convolves with $\mathbi{W}_1$ followed by $\mathbi{W}_3$. 
Because $\mathbi{W}_3$ convolve on the result of ($\mathbi{I} \circledast \mathbi{W}_1 + \mathbi{b}_1$) with an additional circle of zero pixels. 
After merging into one $3\times3$ conv, $\mathbi{I}$ is convolved with an additional padding of  $\mathbi{b}_1$, which is not a constant.
One solution is to add extra modifications $\mathbi{b}_1$ on padding in each training iteration~\cite{ding2021ddb}, which  slows down the training speed.
Instead, we choose to use the opposite order, \ie, employing a $3{\times}3$Conv followed by a $1{\times}1$Conv, In our experiments, this improvement does not bring performance drop while having a faster training. For example, it can decrease the training time from 42h44m to 27h52m (↓~35$\%$) for the RepSR-M16C64 configuration.

\begin{figure}[t]
	\begin{center}
		\includegraphics[width=\linewidth]{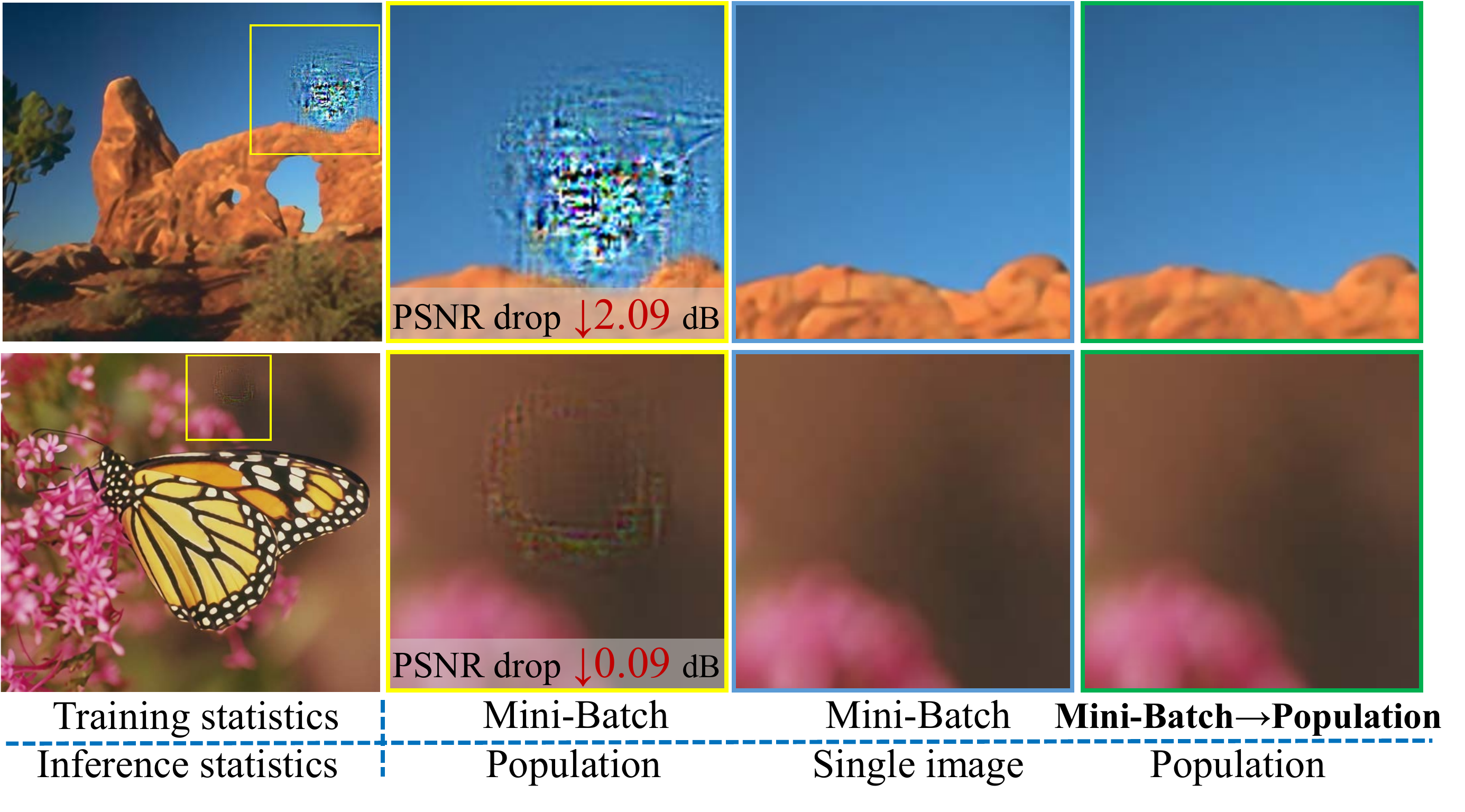}
		\caption{BN tends to produce artifacts in SR, leading to a large performance drop. If we use the single image statistics instead of the population ones, those artifacts can be largely reduced, indicating that the issue is mainly due to the train-test inconsistency. (\textbf{Zoom in for best view})}
		\label{fig:BN_artifacts}
	\end{center}
\end{figure}

\section{Revisiting Batch Normalization in Super-Resolution}
\label{sec:revisit_bn}
In this section, we carefully analyze the causes of  BN artifacts in SR and then propose an effective strategy for removing such artifacts. In such a way, we re-introduce BN into SR under the context of re-paramerization. 

Batch normalization (BN)~\cite{ioffe2015batch} is widely-used in convolution neural networks (CNN) and is effective to ease optimization, thus improving performance.
Given a CNN feature $x$, BN computes the output $y$ by normalizing $x$ using the mean $\mu$ and variance $\sigma^2$:
\begin{equation}
	\label{eq:norm}
	y = \frac{x - \mu}{\sqrt{\sigma^2 + \epsilon}} \times \gamma + \beta,
\end{equation}
where $\gamma$ and $\beta$ are learnable affine parameters and $\epsilon$ is a small value for numerical stability.
During training, the mean and variance are calculated per-channel over the \textit{mini-batches}, while during inference, \textit{population statistics} that estimated from the training set are used for normalization.

However, BN is unfriendly to SR.
It tends to introduce unpleasant BN artifacts~\cite{wang2018esrgan} during inference, as shown in Fig.~\ref{fig:BN_artifacts}. The outputs are messed with unnatural colors, resulting in a large drop of PSNR (peak signal-to-noise ratio).
Sometimes, such artifacts are mild and even imperceptible at the first glance (the bottom, Fig.~\ref{fig:BN_artifacts}), leading to a slight performance drop~\cite{lim2017edsr}.
Therefore, BN is typically removed in SR networks~\cite{lim2017edsr,wang2018esrgan}.
In order to make up for the optimization issue caused by the absence of BN, several alternative techniques, such as residual scaling~\cite{lim2017edsr}, intermediate supervision~\cite{timofte2017ntire} or longer training are applied for training larger networks.

\subsection{Analysis of BN Artifacts}

We start our analysis from an observation -- those BN artifacts are more likely to appear in smooth regions of an image.
Those smooth areas obviously have a very different local variance, indicating that the artifacts are possibly related to incorrect statistics used in BN.
Interestingly, if we normalize features by the statistics of the input image itself instead of the population statistics, the BN artifacts can be largely reduced (3rd column, Fig.~\ref{fig:BN_artifacts}).
This phenomenon implies that the cause of BN artifacts is probably from the \textit{train-test inconsistency}, \ie, the inconsistency between mini-batch statistics and population statistics.

\begin{table}[t]
	\centering
	\tablestyle{8pt}{1.3}
	\begin{tabular}{c|ccccc}
		& Set5 & Set14 & B100 & DIV2K \\
		\shline
		{\textit{w/o} BN} &  {38.21}&  {34.00 } & {32.35} & {36.52}  \\ \hline
		\textit{w/} BN & 38.22 \hl{↑0.01} & 32.44 \textcolor{red}{↓1.56} & 32.19  \textcolor{red}{↓0.16} & 35.87 \textcolor{red}{↓0.66} \\
		\textbf{Proposed} & 38.25  \hl{↑0.04} & 34.07  \hl{↑0.07}& 32.38  \hl{↑0.03} & 36.59  \hl{↑0.07} \\
	\end{tabular}
	\caption{PSNR of EDSR($\times$2) rained on DIV2K. The model with BN achieves good results on the Set5 test dataset, while having a large drop on the other three datesets. With the proposed Frozen BN strategy, it can perform well on all datasets without BN artifacts.
		\label{tab:bn_performance}
	}
\end{table}

In contrast to our common sense, such inconsistency does not happen at the ``dataset'' level. For example, as shown in Tab.~\ref{tab:bn_performance}, the SR model with BN is trained on the DIV2K dataset.
It performs well on the Set5 dataset, while having a large performance drop on its corresponding validation dataset -- DIV2K.
Instead, this inconsistency happens in the \textbf{``patch'' level}.
To be specific, it is the mismatch between statistics of a local patch and the estimated population statistics that results in BN artifacts.

\begin{figure}[t]
	\begin{center}
		\includegraphics[width=\linewidth]{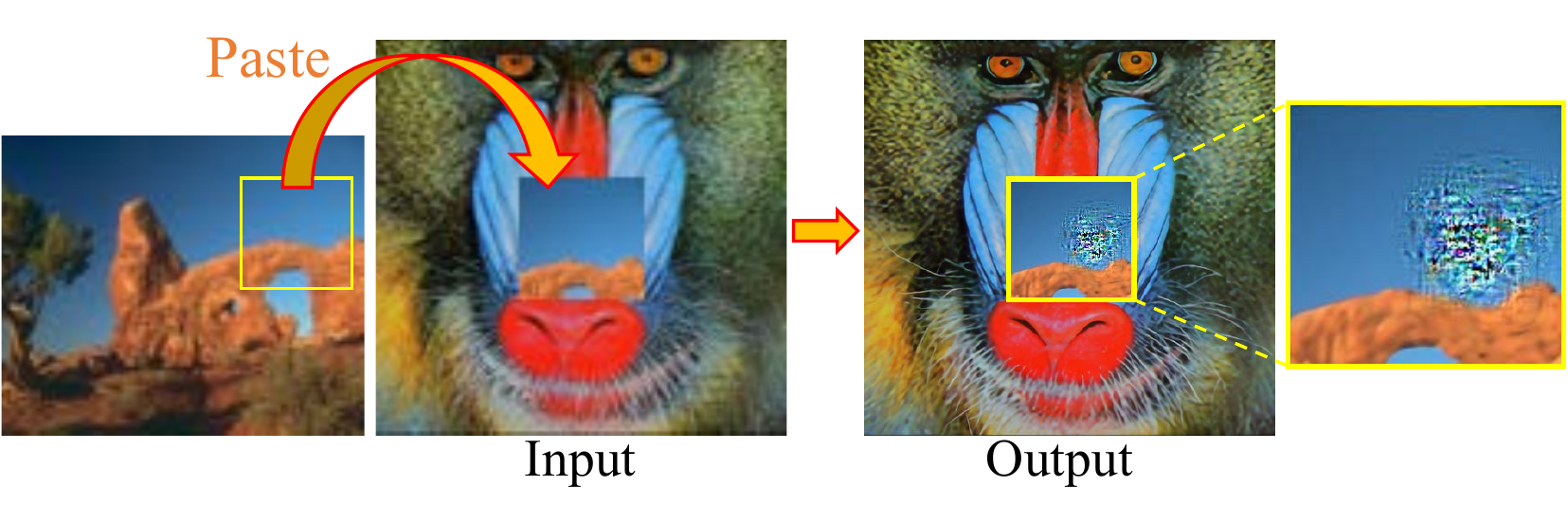}
		\caption{BN artifacts produce at the patch level. We crop and paste the ``problematic'' patch into a normal image. Interestingly, in the output, BN artifacts still exist on the pasted region, while the left is healthy.}
		\label{fig:paste}
	\end{center}
\end{figure}

We conduct an experiment to verify this claim. We crop the LR patch, whose output has serious BN artifacts, from the image in  Fig.~\ref{fig:BN_artifacts}.
Then, we paste this patch on the center of the \textit{baboon} LR image, whose output is apparently good without artifacts, as shown in Fig.~\ref{fig:paste}.
The merged image is then passed through the SR network.
Interestingly, the BN artifacts still exist on the pasted smooth patch, while the left patches are not influenced.
Moreover, the artifact pattern in Fig.~\ref{fig:paste} is almost the same as that in Fig.~\ref{fig:BN_artifacts}.
It seems that those ``problematic'' patches generate abnormal feature statistics as their own characteristics.
As SR networks focus more on local processing~\cite{gu2020interpreting} and keep spatial information from start to end,
the problematic patches will always generate BN artifacts, no matter whether they are alone or together with other patches.

\begin{figure}[tb]
	\begin{center}
		\includegraphics[width=\linewidth]{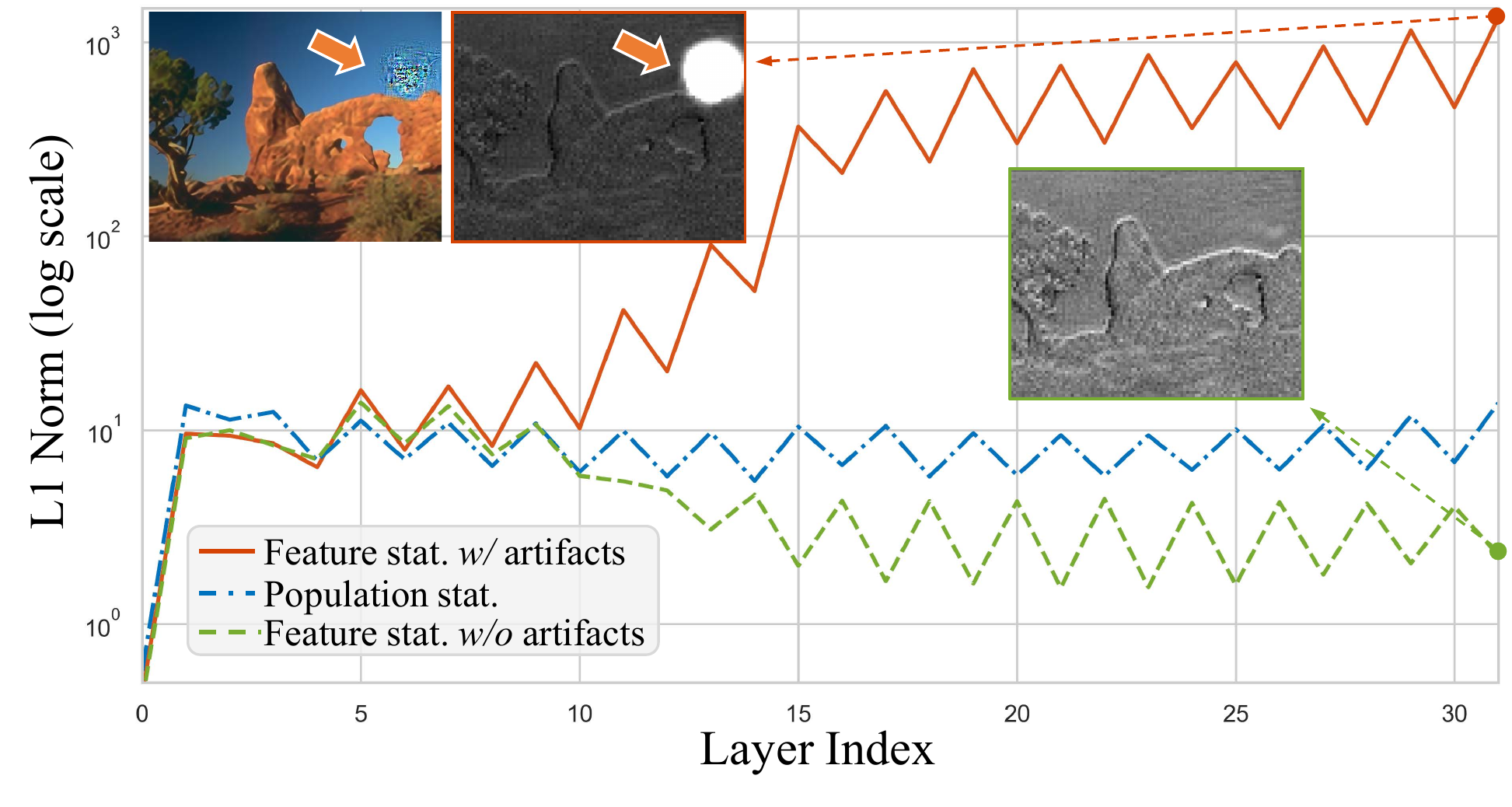}
		\caption{L1 norm curves of \textit{variance}, for feature statistics of outputs with and without artifacts, and the estimated population statistics.}
		\label{fig:bn_std_stat}
	\end{center}
\end{figure}

We then delve into the forming process of BN artifacts.
We examine the L1 norm of statistics in each BN layer.
Fig.~\ref{fig:bn_std_stat} plots feature statistics of outputs with and without artifacts, and the estimated population statistics.
It is observed that the normal feature statistics and population statistics are close and are of the same order of magnitude.
While the feature statistics of abnormal outputs deviate a lot from the population statistics from a certain layer.
These feature statistics obviously cannot be normalized to a normal magnitude with Eq.~\ref{eq:norm}.
As a result, the feature statistics explode and finally generate BN artifacts.
The same phenomenon can be also observed for the mean statistics of each BN layer (see \textit{supplementary}).

The forming process of artifacts also happens at the patch level. We visualize the representative feature maps with and without artifacts in Fig.~\ref{fig:bn_std_stat}.
Only the region of the problematic patch is affected while others are healthy.
During training, the abnormal regions with a deviation between feature statistics and \textit{mini-batch statistics} will be penalized by the loss function, thus being optimized towards deviation elimination.
However, during inference, such abnormal features of problematic patches cannot be normalized to a normal range by the \textit{population statistics}.

In addition, such inconsistency at the ``patch'' level introduces another drawback, as the validation dataset cannot indicate whether a network is healthy or not. Even if a network performs well on all validation datasets, it can still generate BN artifacts on new patches.

\subsection{A Simple Strategy for Removing BN Artifacts}
\label{sec:bn_strategy}

\begin{figure}[t]
	\begin{center}
		\includegraphics[width=\linewidth]{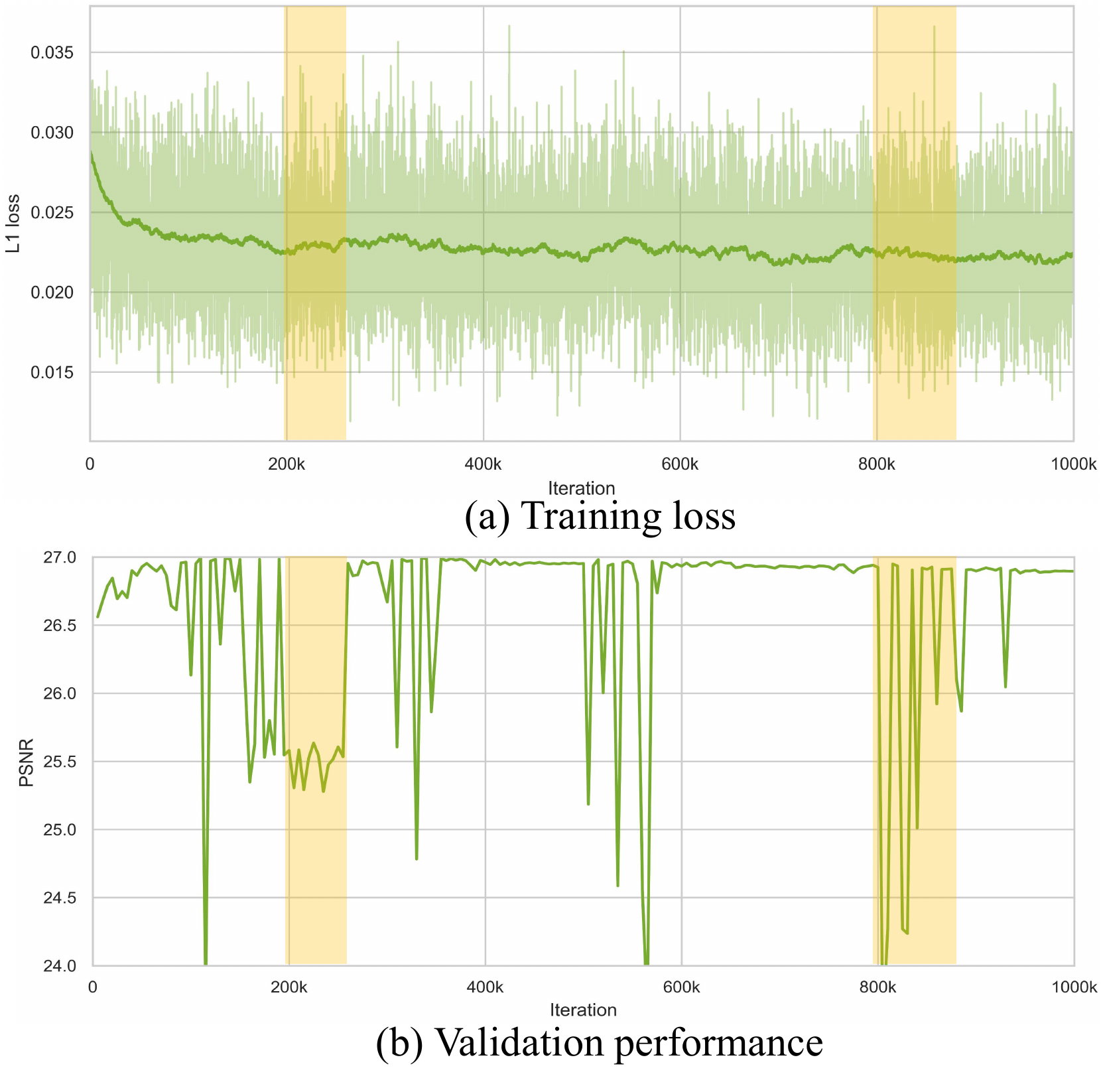}
		\caption{BN artifacts only appear during inference, while the training process is stable and healthy.}
		\label{fig:loss_curve}
	\end{center}
\end{figure}

We then employ a straightforward yet effective strategy for removing BN artifacts.
First, we point out that the BN artifacts only appear during inference, while the training process is stable and healthy.
We examine the training loss and validation performance, as shown in Fig.~\ref{fig:loss_curve}.
It is observed that, 1) the whole training process is steady without abrupt drops. 2) Even in the stages with severely fluctuating validation performance (yellow regions, Fig.~\ref{fig:loss_curve}), the corresponding training loss is still normal and healthy.

Based on our analysis, an intuitive idea of reducing train-test inconsistency is to use population statistics during training.
Therefore, we propose to first train SR networks with \textit{mini-batch statistics}, as normal BN does.
At the later training period, we switch to using \textit{population statistics}.

One may wonder whether this strategy will sacrifice the performance, as the switching process actually changes the optimization behavior of BN layers.
From our experiments, this strategy will not degrade the performance of datasets that originally have high performance. As shown in Tab.~\ref{tab:bn_performance}, both models trained with BN and with our strategy obtain similar and good results on Set 5.
Furthermore, the proposed strategy is capable of removing BN artifacts (last column, Fig.~\ref{fig:BN_artifacts}) and improving performance (Set14, B100, DIV2K results in Tab.~\ref{tab:bn_performance}).
In conclusion, equipped with the proposed strategy, we can enjoy the benefits from faster convergence with BN training, while not being affected by BN artifacts.

This strategy is named FrozenBN in high-level tasks, which is usually for transferring a pretrained model to downstream tasks~\cite{ren2015faster,li2019fully,musgrave2020metric} or for small-batch-size training~\cite{wu2021rethinking}.
We also embrace this strategy in the SR task to reduce the train-test inconsistency, based on our analysis of the cause and properties of the BN issue in SR.

Different from FrozenBN, we do not freeze the affine parameters in BN, as it is not necessary from our experiments.
Adjusting those affine parameters leads to a larger optimization space to remove BN artifacts.
One may also wonder whether we can close the gap between mini-batch and population statistics by enlarging batch size.
Our experiments show that this method cannot effectively remove BN artifacts.
This is mainly due to the large fluctuation brought by the randomness of cropped patches during training, as also reflected by the oscillation of training loss (Fig.~\ref{fig:loss_curve}).

\noindent\textbf{Discussion.}
We have revealed the cause of BN artifacts in SR, and propose an effective strategy to remove such artifacts. Thus, RepSR can enjoy the non-linearity benefits from BN during training, while not being affected by BN artifacts in SR.
This strategy is mainly verified under the context of SR re-parameterization in this work.
It is interesting to investigate whether Forzen BN can further improve performance of general SR task. 
This is beyond the scope of this paper and is left as a future work.

\begin{table*}[t]
	\centering
	\tablestyle{5pt}{1.2}
	\begin{tabular}{l|cccc|ccccc}
		&Model & \#Params & \#FLOPs &\#Conv &Set5 &Set14 &B100 &Urban100 &DIV2K \\
		No.  && (K) & (G) & &PSNR / SSIM &PSNR / SSIM &PSNR / SSIM& PSNR / SSIM &PSNR / SSIM \\
		\shline
		{A0} &{Bicubic} &{—} &{—} &{—} &{28.43 / 0.8113} &{26.00 / 0.7025} &{25.96 / 0.6682}&  {23.14 / 0.6577} & {28.10 / 0.7745} \\
		{A1} & {SRCNN~\cite{dong2014learning}}& {57.00}& {52.7}&  {3}& {30.48 / 0.8628}&{{27.49} / 0.7503}&{\textbf{26.90} / 0.7101} &{\textbf{24.52} / 0.7221}& {29.25 / 0.8090}\\
		{A2} &{ESPCN~\cite{caballero2017real}} & {24.90} & {1.44} & { 3} & { \textbf{30.52} / 0.8697} & { 27.42 / 0.7606} & {26.87 / 0.7216} & {24.39 / 0.7241} & { 29.32 / 0.8120}\\
		{A3} & ECBSR-M4C8~\cite{zhang2021ecbsr} &3.70& 0.21 &6& \textbf{30.52} / 0.8698 &27.43 / 0.7608 &26.89 / 0.7220 &24.41 / 0.7263& 29.35 / 0.8133\\
		\rowcolor{Gray}
		{A4} & \textbf{RepSR-M4C8} &3.70& 0.21 &6& 30.50 / \textbf{0.8706} &\textbf{27.55} / \textbf{0.7634} &26.89 / \textbf{0.7225} &{24.42} / \textbf{0.7274}&  \textbf{29.38} / \textbf{0.8140} \\
		\hline
		{B0} & {FSRCNN~\cite{dong2016accelerating}}&{12.00}&{5.00}&{8}&{30.70 / 0.8657}&{27.59 / 0.7535}&{26.96 / 0.7128}&{24.60 / 0.7258}&{29.36 / 0.8110}\\
		{B1} & SESR-M5C16~\cite{bhardwaj2021collapsible} & 18.32 & 1.05  & 7 &   30.99 / 0.8764 & 27.81 / 0.7624 & \textbf{27.11} / 0.7199 & {24.80} / 0.7389 & 29.65 / 0.8189 \\
		{B2} & ECBSR-M4C16~\cite{zhang2021ecbsr} & 11.90& 0.69& 6& \textbf{31.04} / 0.8805& 27.78 / 0.7693& 27.09 / 0.7283 &24.79 / 0.7422& 29.62 / 0.8197\\
		\rowcolor{Gray}
		{B3} & \textbf{RepSR-M4C16} & 11.90& 0.69& 6& 31.02 / \textbf{0.8817}& \textbf{27.88} / \textbf{0.7715}& 27.10 / \textbf{0.7293} & \textbf{24.82} / \textbf{0.7434}& \textbf{29.66} / \textbf{0.8207}\\
		\hline
		{C0} & {TPSR-NoGAN~\cite{lee2020journey}}&{61.00}&{3.60}&{15}&{31.10 / 0.8779}&{27.95 / 0.7663}&{27.15 / 0.7214}&{24.97 / 0.7456}&{29.77 / 0.8200}\\
		{C1} &{IMDN-RTC~\cite{hui2019lightweight}}&{21.00}&{1.22}&{28}&{31.22 / 0.8844}&{27.92 / 0.7730}&{27.18 / 0.7314}&{24.98 / 0.7504}&{29.76 / 0.8230}\\
		{C2} &SESR-M11C16~\cite{bhardwaj2021collapsible} & 32.14 & 1.85  & 13 &   31.27 / 0.8810 & 27.94 / 0.7660 & 27.20 / 0.7225 & 25.00 / 0.7466 &  29.81 / 0.8221 \\
		{C3} &ECBSR-M10C16~\cite{zhang2021ecbsr}&26.00& 1.50 & 12& \textbf{31.37} / 0.8866 &27.99 / 0.7740 &27.22 / 0.7329& {25.08} / 0.7540& 29.80 / 0.8241\\
		\rowcolor{Gray}
		{C4} &\textbf{RepSR-M10C16}&26.00& 1.50 & 12& 31.36 / \textbf{0.8873} &\textbf{28.09} / \textbf{0.7763} &\textbf{27.23} / \textbf{0.7333}& \textbf{25.10} / \textbf{0.7545}& \textbf{29.83} / \textbf{0.8249}\\
		\hline
		{D0} & {LapSRN~\cite{lai2017lapsrn}}&{813.00}&{149.40}&{27}&{31.54 / 0.8850}&{{28.19} / 0.7720}&{27.32 / 0.7280}&{25.21 / 0.7560}&{29.88 / 0.8250}\\
		{D1} & {EDSR-R5C32~\cite{lim2017edsr}}&{241.80}&{14.15}&{13}&{31.46 / 0.8880}&{28.07 / 0.7760}&{27.27 / 0.7340}&{25.21 / 0.7579}&{29.87 / 0.8256}\\
		{D2} & SESR-M11C32~\cite{bhardwaj2021collapsible} & 114.97 & 6.62  & 13 &   31.54 / 0.8866 & 28.12 / 0.7712 & 27.31 / 0.7277 & 25.31 / 0.7604 & 29.94 / 0.8266 \\
		{D3} & ECBSR-M10C32~\cite{zhang2021ecbsr}& 98.10& 5.65 & 12 &\textbf{31.66} / 0.8911 &28.15 / 0.7776& {27.34} / 0.7363& {25.41} / 0.7653 &29.98 / 0.8281\\
		\rowcolor{Gray}
		{D4} & \textbf{RepSR-M10C32}& 98.10& 5.65 & 12 &\textbf{31.66} / \textbf{0.8920} &\textbf{28.28} / \textbf{0.7811}& \textbf{27.35} / \textbf{0.7372}& \textbf{25.46} / \textbf{0.7674} &\textbf{30.02} / \textbf{0.8291}\\
		\hline
		{E0} & {VDSR~\cite{kim2016vdsr}}&{665.00}&{612.60}&{20}&{31.35 / 0.8838}&{28.02 / 0.7678}&{27.29 / 0.7252}&{25.18 / 0.7525}&{29.82 / 0.8240}\\
		{E1} & {CARN-M~\cite{ahn2018fast}}&{412.00}&{46.10}&{43}&{{31.92} / 0.8903}&{28.42 / 0.7762}&{27.44 / 0.7304}&{25.62 / 0.7694}&{30.10 / 0.8311}\\
		{E2} & {IMDN~\cite{hui2019lightweight}}&{667.40}&{38.41}&{34}&{32.03 / \textbf{0.8966}}&{28.42 / 0.7842}&{\textbf{27.48} / 0.7409}&{\textbf{25.96} / \textbf{0.7843}}&{\textbf{30.22} / 0.8336}\\
		{E3} & ECBSR-M16C64~\cite{zhang2021ecbsr}& 602.90& 34.73 &18 &{31.92} / 0.8946& 28.34 / 0.7817& \textbf{27.48} / 0.7393& 25.81 / 0.7773 &30.15 / 0.8315\\
		\rowcolor{Gray}
		{E4} & \textbf{RepSR-M16C64}& 602.90& 34.73 &18 &\textbf{31.94} / 0.8961& \textbf{28.44} / \textbf{0.7853}& \textbf{27.48} / \textbf{0.7825}& {25.88} / 0.7825 & \textbf{30.22} / \textbf{0.8337} \\
	\end{tabular}
	\caption{\small \textbf{Quantitative results on tiny SR networks (${\times}4$)}.
		Similar to~\cite{zhang2021ecbsr}, PSNR/SSIM are evaluated on Y channel. \#Params, \#FLOPs, and \#Conv represent the number of network parameters, floating-point operations, and convolution layers, respectively. The \#FLOPs is measured under the setting of generating a $1280{\times}720$ SR image. \textbf{Bold} indicates the \textbf{best} within each regime.
		\label{tab:quantative_tiny}
	}
\end{table*}

%% file: sections/4_experiments.tex
\section{Experiments} \label{sec:experiments}
\subsection{Datasets and Implementation}
\label{sec:implementation}
We train our models on the DIV2K training dataset~\cite{agustsson2017ntire} with 800 HR images.
The validation sets are Set5~\cite{bevilacqua2012low}, Set14~\cite{zeyde2010single}, BSD100~\cite{martin2001database}, Urban100~\cite{huang2015single}, and DIV2K validation set~\cite{agustsson2017ntire}. All these datasets are commonly used in SR and licensed for research purposes.
LR images are obtained using the MATLAB bicubic kernel function.

We employ two model sizes: 1) tiny models that have the similar configurations with ECBSR~\cite{zhang2021ecbsr}; 2) large models that are similar to EDSR-M~\cite{lim2017edsr} and SRResNet~\cite{ledig2017srgan}.
The models are trained with L1 loss. The mini-batch size is set to 32.
For tiny models, the LR patch size is set to $64{\times}64$. The initial learning rate is set to $4{\times}10^{-4}$ and decayed with multiple steps.
For the other models, we set the LR patch size to $48{\times}48$. The initial learning rate is set to $2{\times}10^{-4}$ and decayed by the cosine annealing scheduler with four cycles~\cite{loshchilov2016sgdr,wang2019edvr}.
The optimization is performed with Adam optimizer for $1000K$ iterations. The training and analyses are performed with PyTorch on NVIDIA A100 GPUs in an internal cluster.

\subsection{Comparisons to Previous Works}

\noindent\textbf{Results on Tiny Networks}
\noindent We first compare our RepSR on the tiny model size, which is usually designed for mobile devices.
Our aim is to show that our RepSR can achieve good performance on tiny models in a simple and effective way. 
We compare RepSR with representative SR methods, including SRCNN~\cite{dong2014learning}, FSRCNN~\cite{dong2016accelerating}, ESPCN~\cite{caballero2017real}, VDSR~\cite{kim2016vdsr}, LapSRN~\cite{lai2017lapsrn}, CARN~\cite{ahn2018fast}, TPSR~\cite{lee2020journey}, IMDN~\cite{hui2019lightweight}, and recent SR re-parameterization works: ECBSR~\cite{zhang2021ecbsr} and SESR~\cite{bhardwaj2021collapsible}. 
Following the practice in~\cite{zhang2021ecbsr}, we use the notation M$x$C$y$ for different model sizes, denoting $x$ convolutions with $y$ channels (after re-parameterization) in its body structure.

\begin{figure*}[t]
	\begin{center}
		\includegraphics[width=0.95\linewidth]{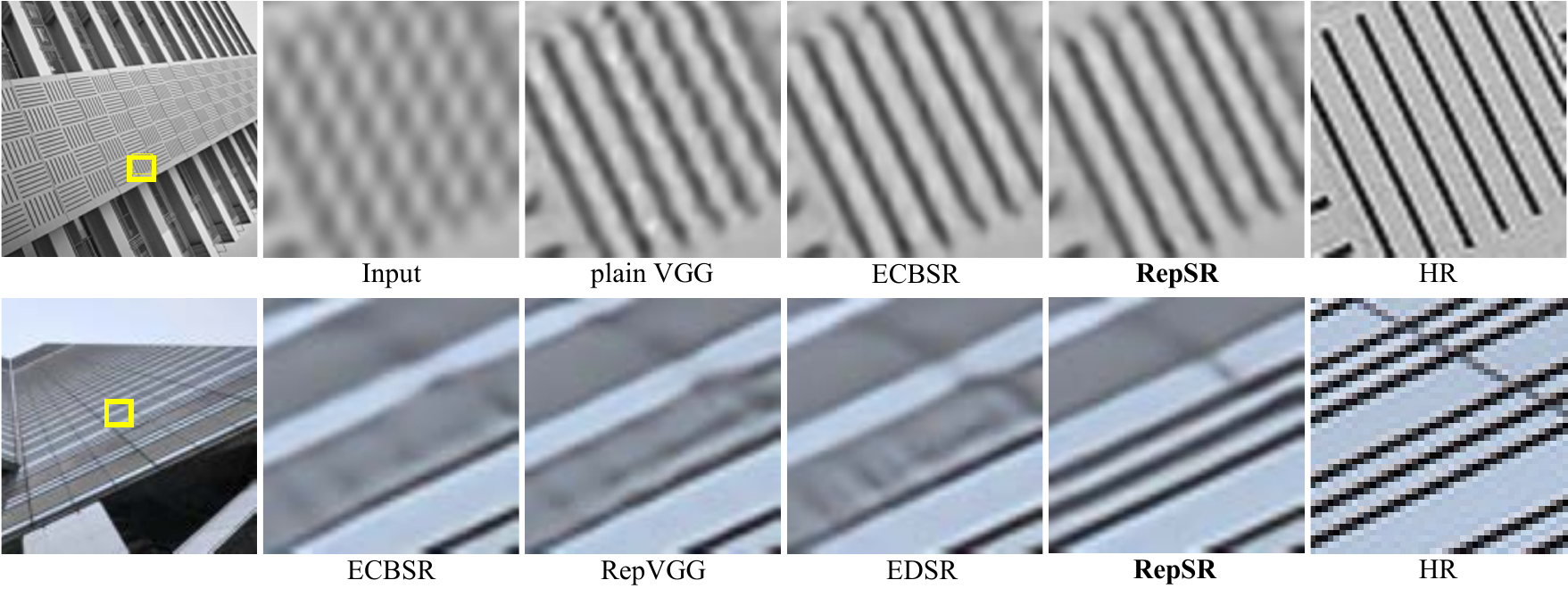}
		\caption{Qualitative comparisons on tiny networks (top) and large networks (bottom).}
		\label{fig:qualitative}
	\end{center}
\end{figure*}

The PSNR/SSIM and model complexity are summarized in Tab.~\ref{tab:quantative_tiny}.
We can draw the following conclusions.
\textbf{1)} Methods with re-parameterization could largely improve performance with fewer parameters and lower computation cost during inference.
\textbf{2)} Our RepSR with a simple structure achieves comparable or better performance on all configurations than ECBSR, while ECBSR requires more complicated structures with manually-designed filters.
\textbf{3)} SESR has a simple block without BN. It has a large performance drop (B1 \textit{v.s.} B3, C2 \textit{v.s.} C4), indicating the importance of BN in re-parameterization.
Qualitative results are shown in Fig.~\ref{fig:qualitative} (top), our RepSR and ECBSR recover better line structures than plain VGG.

We further show that our RepSR has a faster training speed than ECBSR~\cite{zhang2021ecbsr} while having the comparable performance in Tab.~\ref{tab:training_speed}. This is due to the simple structure used in RepSR, while ECBSR adopts complicated pre-defined filters including classical Sobel and Laplacian filters~\cite{sun2008image}.

Plain networks have the advantage in actual running speed and memory consumption~\cite{ding2021repvgg}.
ECBSR~\cite{zhang2021ecbsr} has shown that such plain networks achieve better performance-speed trade-off in mobile devices (Snapdragon 865 and Dimensity 10000+ SOC).
Our RepSR-Tiny has the same inference network as ECBSR and thus can also have fast inference speed and super-resolve frames from 270/540p to 1080p in \textit{real-time} on mobile devices.
\begin{table}[t]
	\vspace{-0.6cm}
	\centering
	\tablestyle{4pt}{1}
	\begin{tabular}{c|ccccc}
		& M4C16 & M10C16 & M10C32 & M16C64 \\
		\shline
		ECBSR~\cite{zhang2021ecbsr} & 9h50m & 19h32m & 24h29m & 55h54m \\
		\textbf{RepSR}  & 4h54m (\hl{↓50$\%$}) & 9h14m (\hl{↓53$\%$}) & 11h38m (\hl{↓52$\%$}) &27h52m (\hl{↓50$\%$}) \\
	\end{tabular}
	\caption{\small Our RepSR can reduce the training time by $\thicksim{50}\%$ than ECBSR, while keeping the comparable performance.
	}
	\label{tab:training_speed}
\end{table}

\begin{table}[t]
	\vspace{-0.6cm}
	\centering
	\tablestyle{6pt}{1.3}
	\begin{tabular}{c|ccccc}
		Method & Set5 & Set14 & B100 & Urban100 & DIV2K \\
		\shline
		plain VGG & 26.26 &24.74  & 25.03  & 22.17 & 26.73 \\
		RepVGG~\cite{ding2021repvgg} & 32.13  & 28.59 & 27.51  & 26.10 & 30.41 \\
		ECBSR~\cite{zhang2021ecbsr} & 31.94   & 28.50  & 27.52   & 25.93  & 30.32 \\
		EDSR~\cite{lim2017edsr} & 32.09    & 28.58  & 27.56    & 26.03   & 30.43  \\
		\textbf{RepSR}~\cite{zhang2021ecbsr} & \textbf{32.17}    & \textbf{28.63}   & \textbf{27.60}    & \textbf{26.19}   & \textbf{30.50}  \\
	\end{tabular}
	\caption{\small Comparisons on different rep-parameterization methods and EDSR~\cite{lim2017edsr}.
	}
	\label{tab:comparison_rep}
\end{table}

\noindent\textbf{Results on Large Networks}

\noindent We then compare our RepSR on large networks.
Instead of huge networks, here, large networks indicates larger model than the above tiny models. We adopt the model size that has a similar running throughput capacity to IMDN~\cite{hui2019imdn} and RFDN~\cite{liu2020residual}. %
We compare our RepSR with SwinIR-Light~\cite{liang2021swinir}, RCAN~\cite{zhang2018rcan}, EDSR~\cite{lim2017edsr}, ECBSR~\cite{zhang2021ecbsr}, IMDN~\cite{hui2019imdn} and RFDN~\cite{liu2020residual}. PSNR are evaluated on the Y channel.
We show the trade-off comparisons between PSNR performance and the throughput in Fig.~\ref{fig:teaser}. 
The throughput is tested on A100 with a batch is 4, fp16 precision and upsampling from $320\times180$ to $1280\times720$. 
Our RepSR is capable of achieving a better balance in the  performance-throughput trade-off.
More quantitative results are provided in the \textit{supplementary}.

We further compare different re-parameterization methods and EDSR~\cite{lim2017edsr} under this setting. PSNR are evaluated on the Y channel.
As shown in Tab.~\ref{tab:comparison_rep},
ECBSR cannot perform well on large networks. Directly applying ECBSR leads to stuckness in low performance. It is mainly due to its absence of BN and too complex hand-designed structures.
Our RepSR can obtain better performance than  RepVGG and ECBSR, and even achieve higher PSNR than the corresponding EDSR with residual blocks.

Qualitative results are shown in Fig.~\ref{fig:qualitative} (bottom), our RepSR can recover clearer lines than other methods, and it even produces better results than EDSR.

\begin{table}[t]
	\centering
	\tablestyle{5pt}{1.1}
	\begin{tabular}{c|c|cc}
		No. & Variants & Set5 & U100  \\
		\shline
		\ding{172} & RepSR & 30.31 & 24.74 \\ \hline
		\ding{173} & remove residual path &29.61 & 24.12 \\
		\ding{174} & residual path with BN & 30.27 & 24.68 \\
		\hline
		\ding{175} & C3-BN + C1-BN & 30.24 & 24.65\\
		\ding{176} & 2${\times}$\{C3-BN-C1-BN\} & 30.30 & 24.74 \\
		\ding{177} & 2${\times}$\{C1-BN-C3\} & 30.31 & 24.75 \\
	\end{tabular}
	\caption{\small Ablation studies on the RepSR block.  C3, C1 denotes $3{\times}3$ and $1{\times}1$ convolutions, respectively.}
	\label{tab:ablation_repsr_block}
\end{table}

\subsection{Ablation Studies}
All the ablation experiments are conducted on the ${\times}4$ model with the large model size. We report PSNR values on RGB channels.

\textbf{Clean residual path.}
We first remove the residual path. As expected, it largely decreases the PSNR (variant \ding{173}, Tab.~\ref{tab:ablation_repsr_block}). If we add BN in the residual path, the performance also has a decline, indicating a clean path is important for SR (variant \ding{174}, Tab.~\ref{tab:ablation_repsr_block})

\textbf{Different choices of other branches.} As shown in Tab.~\ref{tab:ablation_repsr_block}, \textbf{1)} the branches used in RepVGG without expand-and-squeeze conv (variant \ding{175}) brings a PSNR drop. 
\textbf{2)} Adding BN after each convolution  (variant \ding{176}) does not bring further improvements, therefore we only keep one BN between two convolutions for simplicity. 
\textbf{3)} Employing an inverse order of convolution obtains similar results. But the variant \ding{177}  has padding issues during each training iteration. Thus, we adopt the C3-BN-C1 configuration. In our experiments, this improvement does not bring performance drop while having a faster training. For example, it can decrease the training time from 42h44m to 27h52m (↓~35$\%$) for the RepSR-M16C64 configuration.

\textbf{Width multiplier.} We adjust the width multiplier in the expand-and-squeeze convolutions.
It is observed from Tab.~\ref{tab:ablation_wdith} that when switching the width multiplier from one to two, there is an improvement. Only a slight improvement can be obtained if the width multiplier further increases.
Thus, we set width multiplier to 2 in our experiments.

\textbf{Number of branches.}
We change the number of non-residual branches in RepSR. As we can see, one branch leads to an apparent performance drop, while the performance saturates if we further increase the branch number.
Thus, we set the number of branches to 2.

\begin{table}[t]
	\centering
	\tablestyle{5pt}{1.1}
	\begin{tabular}{c|cccc}
		\# Width multiplier & 1 & 2 & 3 & 4 \\
		\shline
		Set5 & 30.27 & 30.31 & 30.30 & 30.33 \\
		Urban100 & 24.71 & 24.74 & 24.76 & 24.75 \\
	\end{tabular}
	\caption{\small Ablation studies on width multiplier.
	}
	\label{tab:ablation_wdith}
\end{table}

\begin{table}[t]
	\centering
	\tablestyle{5pt}{1.1}
	\begin{tabular}{c|cccc}
		\# Branch & 1 & 2 & 3 & 4 \\
		\shline
		Set5 & 30.25 & 30.31 & 30.29 & 30.31 \\
		Urban100 & 24.71 & 24.74 & 24.75 & 24.75 \\
	\end{tabular}
	\caption{\small Ablation studies on the number of branches.
	}
	\label{tab:ablation_depth}
\end{table}

\subsection{Limitations}

Although RepSR can successfully train VGG-style SR networks and achieve better performance-throughput balance, it still has several limitations.
1) RepSR only works for training powerful plain networks, and cannot further improve the peformance for other strong blocks, such as EDSR and DenseNet. 
Such an observation is also discussed in~\cite{ding2021repvgg}.
2) Empirically, we find that very deep plain networks (\eg, more than 100 layers) are still challenging for RepSR, whose performance cannot reach its EDSR counterpart.